# Deep Reinforcement Learning for Mobile Robot Path Planning

Hao Liu[1,*], Yi Shen[2], Shuangjiang Yu[3], Zijun Gao[4], Tong Wu[5]

[1] Northeastern University, Shenyang, China
[2] University of Michigan, Ann Arbor, United States
[3] Northeastern University, Shenyang, China
[4] Northeastern University, Boston, MA, United States
[5] University of Washington, Seattle, USA
*Correspondence Author, liuhao@stumail.neu.edu.cn

**Abstract:** Path planning is an important problem with the the applications in many aspects, such as video games, robotics etc. This paper proposes a novel method to address the problem of Deep Reinforcement Learning (DRL) based path planning for a mobile robot. We design DRL-based algorithms, including reward functions, and parameter optimization, to avoid time-consuming work in a 2D environment. We also designed an Two-way search hybrid A* algorithm to improve the quality of local path planning. We transferred the designed algorithm to a simple embedded environment to test the computational load of the algorithm when running on a mobile robot. Experiments show that when deployed on a robot platform, the DRL-based algorithm in this article can achieve better planning results and consume less computing resources.

**Keywords:** Mobile robot; Deep reinforcement learning; Path planning; Navigation.

## 1. INTRODUCTION

Deep reinforcement learning is widely used in News Recommendation [1], face recognition [2], Image detection [3], task decision-making [4],etc. In terms of robots, deep reinforcement learning is also widely used in robot localization [5], perception[6], decision-making [7], planning [8], hardware [9], etc. In the field of exploring intelligent systems, deep reinforcement learning (DRL) has become a powerful tool for achieving complex decision-making and path planning tasks [10]. The core advantage of DRL is its ability to learn optimal policies through interaction with the environment, without the need for pre-defined rules or models [11]. This characteristic makes it show great potential and application value in fields such as autonomous vehicles, robot navigation, and path planning in complex environments. With the rapid development of deep learning technology, DRL has become more and more widely used in path planning, providing new perspectives and methods for intelligent system design.

The path planning method of deep reinforcement learning mainly optimizes the path selection from the starting point to the end point through the interaction between the learning agent and the environment to maximize the goal. In this process, DRL combines the powerful feature extraction capabilities of deep learning with the decision-making mechanism of reinforcement learning, enabling the system to effectively learn and adapt in complex and changing environments. Through trial and error, the agent learns how to develop the best course of action when encountering obstacles, dynamically changing environmental conditions, and unknown factors [12].

However, the application of DRL in path planning also faces many challenges. For example, how to balance the contradiction between exploration and exploitation, and how to design an effective reward mechanism to guide the agent's learning process. In addition, the processing of high-dimensional state space and continuous action space is also an important issue that DRL needs to solve. In recent years, researchers have made significant progress in these areas by introducing advanced algorithms such as Deep Deterministic Policy Gradients (DDPG), Double Q-learning, and Soft Action Policy Gradient (SAC) [13]. However, in the actual application process, the deep reinforcement learning model training process consumes too much computing resources, and there may be many problems in the deployment process of the embedded platform [14].

In addition, the trajectories generated based on deep reinforcement learning usually do not conform to the actual kinematic constraints of the robot. Due to the incompleteness of the movement of most robots, many problems will occur during the actual control process of the robot. The generated trajectory cannot be tracked, resulting in the final The robot planning failed and entered the shock stage.

Based on the Soft-Actor-Critic (SAC) algorithm of the maximum entropy framework, this paper proposes a bidirectional parallel SAC (Parallel SAC, PSAC) training algorithm to solve the problem of incomplete robot path planning algorithms in actual complex scenarios. question. At the same time, kinematics is combined with the hybrid A* algorithm, and the sub-method is used as a local path planner to track the generated trajectory. The final planning results were verified in the actual robot platform, and the experiments showed the effectiveness and stability of the above method.

## 2. RELATED WORK

In the field of robotic path planning, researchers are continuously advancing techniques for more efficient and precise control, employing a variety of computational methodologies. The latest studies in this domain highlight significant interdisciplinary applications and theoretical improvements that enhance robotic capabilities.

Li et al. [15] explored the use of the LightGBM algorithm for credit assessment, showcasing its potential for optimizing decision-making processes in robotics. Zhang et al. [16] focused on hardware-based real-time workload forensics, which can enhance robotic systems' response times and operational accuracy. Liu et al. [17] employed a novel graph neural network approach for rumor detection, which can be adapted to network-based path planning in robotics.Tong et al. [18] enhanced visual navigation for robots through adaptive cost volume representation in stereo matching, essential for navigating complex environments. Su et al. [19] and Zhang et al. [20] discussed the strategic implications of Monte Carlo Tree Search algorithms and large language models, respectively, which are crucial for developing advanced decision-making frameworks in robotics.Huang et al. [21], [22] explored federated learning techniques to improve data privacy and model robustness in distributed robotic systems. Ru et al. [23] contributed insights into underwater vehicle trajectory planning, providing valuable methods for navigating challenging aquatic environments. Li et al. [24] developed a new planning algorithm for unicycle-type mobile robots, focusing on precise trajectory tracking.Huang and Yang [25] introduced symmetric contrastive learning for fault detection in traffic sensor data, a method that can be applied to robotic system diagnostics to ensure reliable operations.

Path planning is one of the important directions in robot related fields, aiming to find a safe path for the robot to pass between the starting point and the target.Generally speaking, path planning is divided into global path planning and local path planning, with the difference that all environmental information in the current environment can be accessed [26]. In global path planning algorithms, all environmental information is known by default, while in local path planning, robots usually can only obtain environmental information about their own surroundings based on their perception ability [27], and are unaware of the environment outside their perception range. Therefore, global path planning is often applied in closed environments that have already been modeled, while local path planning is used in related directions such as autonomous navigation [28]. Currently in engineering applications, this is often used in combination [29].

Deep reinforcement learning (DRL) initially made its mark in the discrete control of mobile robots for obstacle avoidance [30]. Following this, Pfeiffer et al. [31] utilized the outcomes from Dijkstra and Dynamic Window Approach (DWA) as training labels within convolutional neural networks (CNNs) for developing a path planner. Concurrently, Tai et al. [32] introduced a comprehensive strategy for obstacle avoidance in mobile robots using deep Q-learning, and further extended their work to develop a map-free indoor navigation system leveraging sparse laser radar data and DRL techniques [33]. Meanwhile, Faust et al. [34] combined Probabilistic Roadmaps (PRM) with reinforcement learning to facilitate long-range navigation in robots, a technique also adopted by Francis et al. [35] who integrated AutoRL with PRM for enhanced indoor navigation.

Expanding upon these concepts, Zeng et al. [36] enhanced robot navigation in dynamic environments through a novel approach using an asynchronous advantage Actor-Critic (A3C) algorithm modified with jump point search (JPS-IA3C). This method calculates an abstract path using JPS+ as the global planner while learning local motion control policies via the improved A3C algorithm, thus harmonizing local and global planning efforts for robotic pathfinding. Iyer et al. [37] explored vehicle collision avoidance using a meta-learning strategy known as CARML, implemented on a 2D navigation system equipped with a lidar sensor, and benchmarked against a traditional TD3 solution.

In the realm of aerial robotics, Sánchez-López et al. [38] introduced a real-time 3D path planning system using a probabilistic graph that samples permissible spaces, ignoring existing obstacles. This graph is then navigated using an A* search algorithm integrated with an artificial field map to derive an optimal collision-free route. Additionally, Kim et al. [39] developed a motion planning algorithm that incorporates Twin Delayed Deep Deterministic Policy Gradient (TD3) with hindsight experience replay[40], [41], improving the efficiency of path planning to yield smoother and shorter routes than those generated by PRM systems[42], [43].

## 3. MODEL AND ALGORITHM

### 3.1 Vehicle kinematic model

The vehicle kinematic model is relatively complex, and some approximations are made during the modeling process to facilitate subsequent derivation and calculations.

The Ackermann turn model can usually be expressed in the following figure:

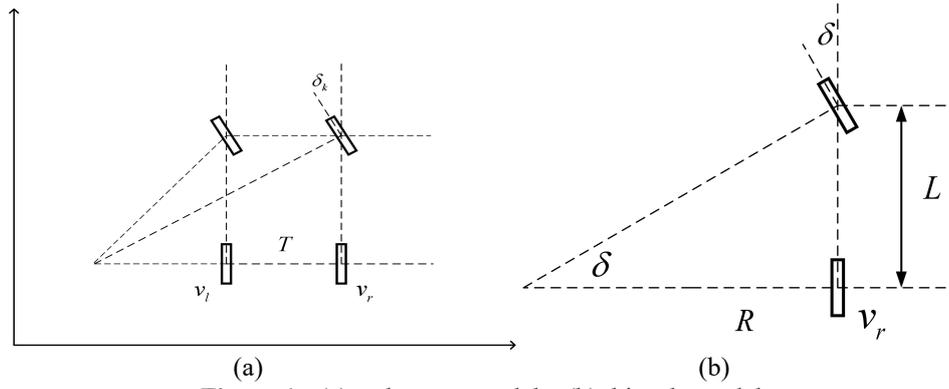

**Figure 1:** (a) ackerman model, (b) bicycle model

When turning the vehicle, it is necessary to make the instantaneous turning centers of the left and right front wheels coincide with the same point on the straight line where the rear wheels are located. Therefore, in practical operation, the turning angle of the inner wheel is slightly larger than that of the outer wheel, so that the two wheels will take on a trapezoidal shape instead of a parallelogram.

For easier analysis, vehicle models are usually simplified. One simplification method is to use the so-called "bicycle model", which involves imagining a tire at the center of the front wheel and the center of the rear wheel to represent the motion state of the front and rear wheels. This simplified model can make it easier to study the motion characteristics and performance of vehicles.

The control variables are usually selected as rear wheel speed $v$ and front wheel angle $\delta$, while the state variables are selected as coordinates in the world coordinate system $x$, $y$ and heading angle $\theta$. Therefore, the kinematic model of Ackermann steering can be represented as

$$\begin{cases} x' = v\cos(\theta) \\ y' = v\sin(\theta) \\ \theta' = v\tan(\delta)/L \end{cases} \quad (1)$$

$$\begin{cases} x_{t+1} = x_t + v_t \cos(\theta_t) d_t \\ y_{t+1} = y_t + v_t \sin(\theta_t) d_t \\ \theta_{t+1} = \theta_t + \omega' d_t \end{cases} \quad (2)$$

**3.2 Soft Actor-Critic Algorithms**

The Soft Actor-Critic (SAC) algorithm is an efficient, entropy-based deep reinforcement learning method that finds optimal policies in a continuous action space. This approach mainly focuses on improving the performance and exploration efficiency of the policy simultaneously, by introducing entropy as an additional reward signal to encourage exploration, thereby making the learning process more robust and effective.

The standard reinforcement learning objective is the expected sum of rewards $\sum_t E(s_t, a_t) \sim \rho_\pi [r(s_t, a_t)]$ and our goal is to learn a policy $\pi(a_t | s_t)$ that maximizes that objective. The maximum entropy objective generalizes the standard objective by augmenting it with an entropy term, such that the optimal policy additionally aims to maximize its entropy at each visited state:

$$\pi^* = \arg\max_\pi \sum_t E(s_t, a_t) \sim \rho_\pi [r(s_t, a_t) + \alpha N(\pi(\cdot | s_t))] \quad (3)$$

where $\alpha$ is the temperature parameter that determines the relative importance of the entropy term versus the reward, and thus controls the stochasticity of the optimal policy.

The reward function is the reward and punishment value for the robot to take a certain action in a certain state. This paper designs a continuous reward function, which is more conducive to the convergence of the model than the sparse reward, allowing the robot to acquire obstacle avoidance and navigation capabilities more quickly. The reward function consists of two items:

$$reward = step + final\_goal \tag{4}$$

In the formula: $step$ is the reward obtained by the robot at each step, which is determined by the difference between the distance of the robot from the target point in the previous step and the distance of the current step from the target point. $final\_goal$ is the reward at the end of the robot's movement. If the robot collides, $final\_goal$ is -20; if it reaches the target point successfully, the value is +25; otherwise, there is neither a collision nor the target point reached within a certain movement time. Then the motion state ends and the value is 0.

**3.3 Hybrid A***

The Hybrid A * algorithm is a path planning algorithm proposed by Dmitri Dolgov, Sebastian Thrun, Michael Montemerlo, and others from Stanford University in 2010. This algorithm solves the problem that traditional A * algorithms cannot meet the kinematic characteristics of robots by combining path planning with the kinematic characteristics of robots. Compared with the traditional A * algorithm, the advantage of the hybrid A * algorithm is that it can be applied to the path planning problem of Ackermann model robots and commonly used vehicles, and has shown good results in lateral parking and reverse parking. However, the drawback of this algorithm is that it cannot guarantee the optimality and completeness of path planning.

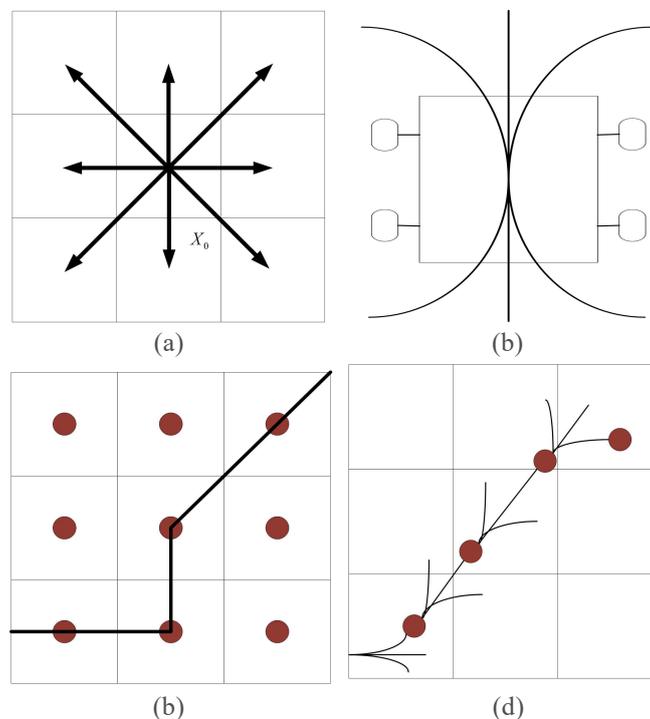

**Figure 2:** (a) Expansion in eight directions , (b) Actual way of moving unmanned vehicles, (c) node in Expansion in eight directions, (d) Hybrid A* algorithm expansion node

Real time path planning requires high computational speed. In many cases, such as a robot moving at high speed in an unknown environment, the trajectory needs to be continuously regenerated in a short period of time to avoid sudden threats. In general, path planning needs to be re planned mostly because the original path quality is poor. Due to the lack of optimality in the hybrid A * algorithm, the first path searched is not the optimal path. In the process of path search, on the one hand, we adopt the method of opposite search to improve the completeness of the algorithm. On the other hand, this algorithm does not end the search process when a path is found, but continuously generates new paths during the search process. By generating multiple paths at once, we comprehensively evaluate the quality of the multiple paths and obtain the best quality path.

**4. EXPERIMENTS**

**4.1 SAC Algorithms for path planning**

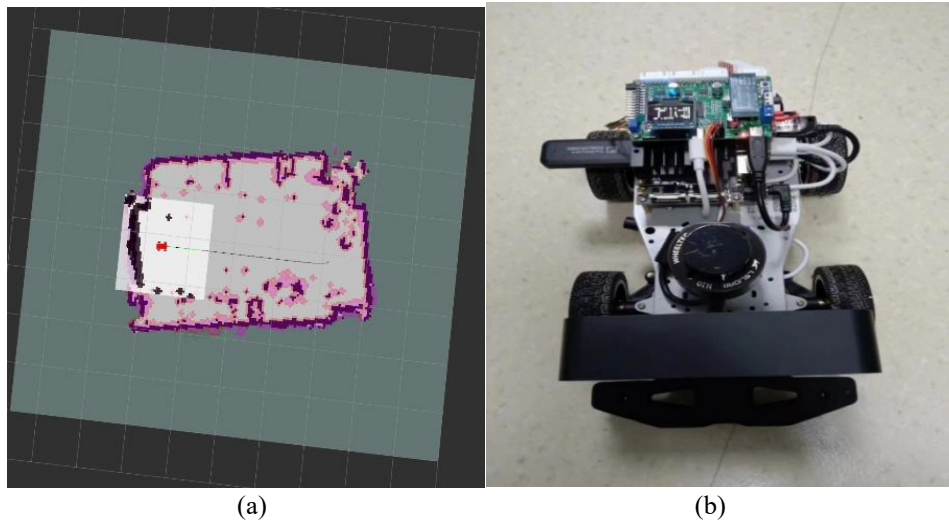
(a)                          (b)
**Figure 4: (a)** Unmanned vehicle rviz simulation, (b) Unmanned vehicle physical display

The unmanned vehicle is an Ackerman model unmanned vehicle, its main control is Jetson nano, and the CPU is ARM Cortex-A57 64-bit@1.43GHz. The lidar is Leishen N10P lidar, with a measurement radius of 25 (m) and a sampling frequency of 5400 (Hz). Experiments show that the SAC algorithm can run well on this platform and successfully generate global paths.

**4.2 Two-way search Hybrid A***

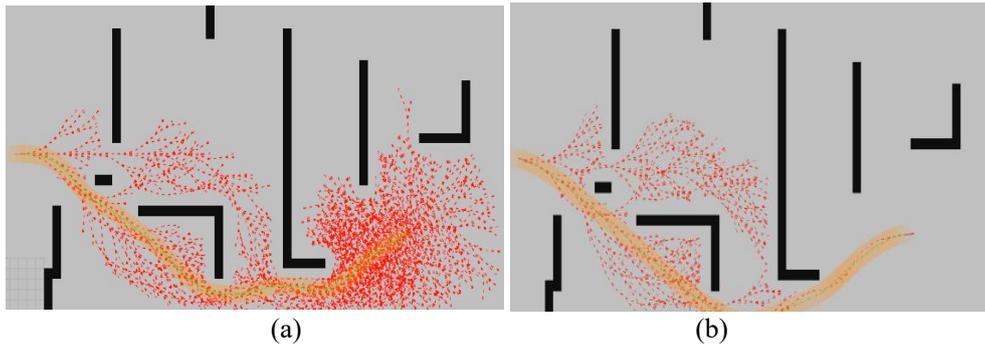
(a)                          (b)
**Figure 3:** Comparison between the algorithm (a) Hybrid A* , (b)Two-way search Hybrid A*

Comparing the significant differences in completeness between the one-way hybrid A* algorithm and the two-way A* algorithm designed in this article, it can be seen from the figure that when searching in one direction, the algorithm has been expanded multiple times, but still cannot get close. End point, this not only leads to a significant reduction in overall algorithm efficiency, but also has a fatal impact on the completeness of the algorithm. The bidirectional hybrid A* algorithm can quickly expand from the local compressed environment through end point expansion, thereby quickly realizing path planning. And it can be seen in table 4.1 that the Two-way search Hybrid A* algorithm has high computational efficiency and few expansion points.

**Table 1**: Comparative data

| Margin | Hybrid A* | Two-way search Hybrid A* |
|---|---|---|
| Expand point | 3152 | 845 |
| Searching time | 45 mm | 12 mm |

## 5. CONCLUSION

This study successfully demonstrates the application of a DRL-based algorithm and a two-way search Hybrid A* algorithm in mobile robot path planning. The innovations presented in this research effectively address several challenges commonly associated with path planning, such as high computational load and the need for rapid, real-time path adjustments in dynamically changing environments. Our results show that integrating DRL with efficient path planning algorithms can significantly enhance robotic navigation systems, offering a scalable solution for operating in complex environments.The combined use of DRL and the two-way search Hybrid A* algorithm enables a more robust navigation system capable of adapting to various environmental changes and obstacles while minimizing the computational resources required. This integration not only improves the efficiency of the path planning process but also ensures higher reliability and better performance of the robotic systems in real-world applications.